\PassOptionsToPackage{table,dvipsnames}{xcolor}
\PassOptionsToPackage{hyphens}{url}
\documentclass[10pt,twocolumn,letterpaper]{article}

\IfFileExists{cvpr.sty}{%
  \usepackage[pagenumbers]{cvpr}%
}{%
  \usepackage[pagenumbers]{arxiv/cvpr}%
}

\usepackage{multirow}
\usepackage{makecell}
\usepackage{tabularx}

\newcolumntype{Y}{>{\centering\arraybackslash}X}

\definecolor{cvprblue}{rgb}{0.21,0.49,0.74}
\usepackage[pagebackref,breaklinks,colorlinks,allcolors=cvprblue]{hyperref}

\def\paperID{}
\def\confName{CVPR}
\def\confYear{2026}

\newcommand{\inputsection}[1]{%
  \IfFileExists{sections/#1}{\input{sections/#1}}{\input{arxiv/sections/#1}}%
}
\newcommand{\printreferences}{%
  \IfFileExists{main.bbl}{%

  }{%
    \IfFileExists{arxiv/main.bbl}{%
    }{%
      \IfFileExists{ieeenat_fullname.bst}{%
        \bibliographystyle{ieeenat_fullname}
      }{%
        \bibliographystyle{arxiv/ieeenat_fullname}
      }%
      \IfFileExists{references.bib}{%
        \bibliography{references}
      }{%
        \bibliography{arxiv/references}
      }%
    }%
  }%
}
\IfFileExists{Figure_1.pdf}{\graphicspath{{./}}}{\graphicspath{{arxiv/}}}

\title{Engram-E2VID: Reference-Based Event-to-Video Reconstruction via Generative Activation of Appearance Engrams}

\author{
Feiyu Ji\textsuperscript{1} \quad
Xiang Li\textsuperscript{1} \quad
Hao Ma\textsuperscript{1} \quad
Tianxiang Huang\textsuperscript{1} \quad
Qingxin Lu\textsuperscript{1}\\
Mengqi Ji\textsuperscript{2} \quad
Lei Han\textsuperscript{3} \quad
Xiaokang Yang\textsuperscript{1} \quad
Xiaoyun Yuan\textsuperscript{1}\\
\textsuperscript{1}MoE Key Lab of Artificial Intelligence, Institute of AI, School of Computer Science,\\
Shanghai Jiao Tong University\\
\textsuperscript{2}Institute of Artificial Intelligence, Beihang University\\
\textsuperscript{3}DISCOVER Robotics\\
{\tt\small yuanxiaoyun@sjtu.edu.cn}
}

\begin{document}

\maketitle
\begin{abstract}
Reference-based event-to-video reconstruction aims to recover target RGB frames from a reference frame and the event stream captured over the reference-to-target interval. Although events provide fine-grained temporal cues, they encode sparse and asynchronous log-intensity changes rather than absolute appearance, making faithful reconstruction intrinsically challenging. The central challenge lies in associating event-derived target-time structures with relevant appearance information from the reference frame, especially under complex motion and long temporal intervals. In this work, we propose Engram-E2VID, a structure-guided framework that reconstructs target frames through the generative activation of appearance engrams. Specifically, the reference frame is encoded into token-space appearance engrams, while the event stream and reference context are transformed into a target-time motion-structure scaffold that captures motion boundaries and event-induced structural changes. Within a one-step diffusion backbone, scaffold-derived structural tokens progressively interact with and activate relevant appearance engrams across layers. This token-space association allows target structures to access reference appearance without relying on direct pixel-wise correspondence, while the diffusion prior complements uncertain or newly revealed regions. Across three benchmarks, Engram-E2VID improves PSNR by up to 3.29 dB and reduces LPIPS by up to 0.08 over the strongest same-input baseline, while degrading more slowly as the reconstruction interval increases.
\end{abstract}

\section{Introduction}

Event cameras provide an alternative paradigm for sensing dynamic scenes. Instead of capturing dense intensity images at a fixed frame rate, they asynchronously record per-pixel log-intensity changes, offering microsecond-level temporal resolution and high dynamic range~\cite{9138762}. However, events encode only relative intensity changes and do not directly preserve absolute intensity, color, or texture, making event-only video reconstruction inherently ill-posed~\cite{scheerlinck2020fast, 8946715}. In this work, we study reference-based event-to-video reconstruction, which aims to recover RGB frames at specified target timestamps from a reference frame and the subsequent event stream. The reference frame provides appearance information and scene context, whereas events offer fine-grained temporal evidence of scene changes.

Despite this complementarity, associating reference appearance with target-time structures inferred from sparse and asynchronous events remains challenging~\cite{zhou2025bridge}. As the temporal interval increases, event noise and motion ambiguity accumulate~\cite{10713104}, while large displacements and substantial structural changes further weaken the correspondence between the reference and target frames. Many existing reference-based methods estimate motion fields from events and warp the reference frame to the target timestamp~\cite{9950520, zhu2024video}. Although effective under locally coherent motion, warping alone cannot reliably reconstruct newly revealed or substantially altered regions. Recent diffusion-based approaches introduce generative priors to recover uncertain appearance details, but typically incorporate events through auxiliary conditioning branches~\cite{Chen_2025_CVPR}. Such conditioning lacks an explicit mechanism for associating event-derived target-time structures with reference-frame appearance cues, potentially leading to structurally inconsistent or hallucinated content.

To address this limitation, we propose Engram-E2VID, a structure-guided generative framework for reference-based event-to-video reconstruction. Our key design is an interaction interface that assigns distinct roles to the two observations: a motion-structure scaffold represents target-time motion and structure, while reference-derived latent tokens serve as appearance engrams. Within a one-step diffusion backbone, scaffold-derived structural tokens repeatedly interact with and activate relevant appearance engrams across layers. Because this association is established in token space rather than through direct pixel transfer, target structures can access relevant reference appearance despite substantial spatial displacement. The diffusion prior further complements appearance information in uncertain or newly revealed regions, reducing structural distortions and visually implausible details.

We evaluate Engram-E2VID on multiple synchronized event-frame datasets acquired using different event cameras. Across these benchmarks, Engram-E2VID consistently yields more accurate reconstructions with sharper object boundaries and fewer visual artifacts. In summary, our main contributions are as follows:

\begin{itemize}
\item We introduce a target-time motion-structure scaffold that integrates events with reference context to provide explicit structural guidance for reconstruction.

\item We establish a structure-to-engram interaction interface in which scaffold-derived structural tokens activate relevant appearance engrams within a one-step diffusion backbone.

\item Extensive experiments across multiple benchmarks demonstrate substantial improvements in visual quality and structural fidelity, together with slower degradation as the reconstruction interval increases.
\end{itemize}

\section{Related Work}

\subsection{Event-Based Video Reconstruction}

Event-based video reconstruction aims to recover dense intensity frames from asynchronous event streams. Existing approaches can be broadly divided into event-only and frame-assisted reconstruction. Event-only methods learn to infer intensity sequences by modeling the spatiotemporal dynamics encoded in event streams~\cite{Rebecq_2019_CVPR,9337171,9710343}. Although these methods can reconstruct temporally dense videos, events record only relative log-intensity changes and do not preserve absolute color or texture, making faithful appearance recovery inherently ambiguous.

Frame-assisted methods reduce this ambiguity by combining events with intensity observations~\cite{9252186,Yang_2023_CVPR}. Early approaches integrate event measurements under explicit imaging models, whereas recent reference-based methods estimate event-guided motion for reference-frame warping and inpainting~\cite{zhu2024video}, or combine continuous motion modeling with learned feature fusion~\cite{Wang_2025_CVPR}. These methods associate target content with the reference primarily through motion estimation and spatially organized feature transfer. Under complex motion, occlusion, and substantial reference-target displacement, however, reliable association between target-time structures and reference appearance remains challenging.

\subsection{Diffusion Models with Event Guidance}

Diffusion priors have recently been incorporated into event-based frame reconstruction to recover visual information that is weakly constrained by sparse event measurements~\cite{NEURIPS2020_4c5bcfec,Rombach_2022_CVPR}. Existing methods adopt different interfaces between events and diffusion models. Event-Diffusion refines preliminary event reconstructions through diffusion-based restoration~\cite{liang2023eventdiffusion}, while temporal residual guided diffusion reconstructs inter-frame residuals using temporal and frequency-domain event cues~\cite{zhu2024temporal}. RE-VDM adapts a pretrained video diffusion model to event-guided frame interpolation~\cite{Chen_2025_CVPR}. More recently, DESSERT aligns event representations with inter-frame residual latents and performs event-conditioned residual diffusion~\cite{kong2025dessert}.

These approaches employ events as refinement cues, diffusion conditions, or residual priors. Engram-E2VID instead focuses on structuring the interaction between target-time evidence and reference appearance. It converts events and reference context into a motion-structure scaffold and organizes reference appearance as token-space engrams, allowing scaffold tokens to activate relevant engrams throughout a one-step diffusion backbone. This design avoids rigid pixel-wise correspondence while using the diffusion prior to resolve event ambiguity and suppress artifacts.

\section{Method}
\subsection{Framework Overview}
\begin{figure*}[t]
\centering
\includegraphics[width=\textwidth]{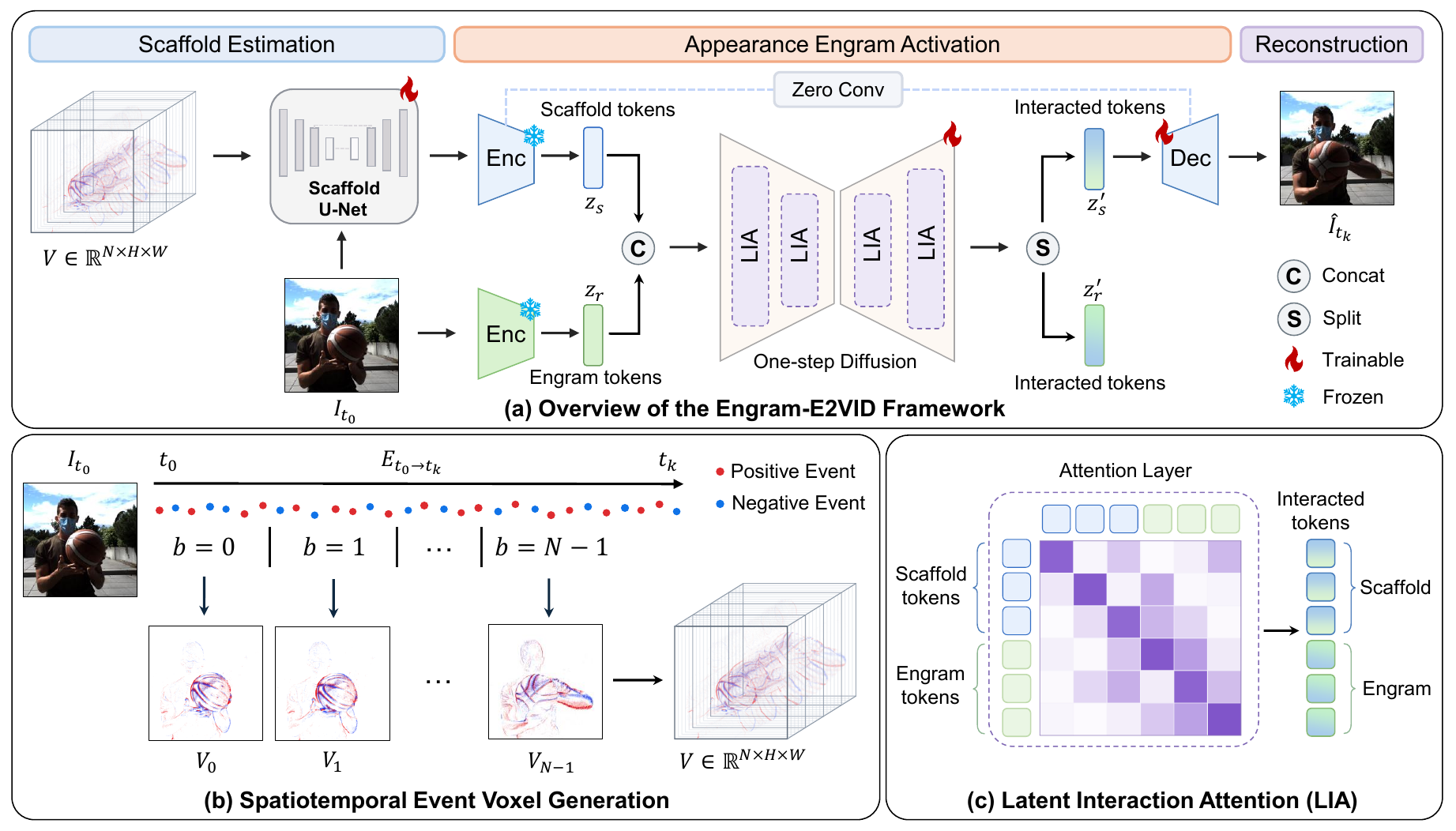}
\caption{
(a) Overview of the Engram-E2VID framework. The Scaffold U-Net estimates a target-time motion-structure scaffold from $I_{t_0}$ and $V$, which interacts with the reference latent through LIA modules within a one-step diffusion backbone for target-frame reconstruction.
(b) Spatiotemporal event voxel generation. Events in $E_{t_0\rightarrow t_k}$ are divided into $N$ temporal bins to form $V\in\mathbb{R}^{N\times H\times W}$.
(c) Latent Interaction Attention. LIA rearranges the scaffold and reference latents into a joint token sequence for latent interaction and then restores their spatial organization.
}
\label{fig:framework}
\end{figure*}

In this section, we present Engram-E2VID, a reference-based event-to-video reconstruction framework based on the generative activation of appearance engrams. As illustrated in Fig.~\ref{fig:framework}(a), Engram-E2VID consists of two key components. First, Structural Scaffold Estimation converts the temporally binned events and reference frame into a scaffold image containing motion-structure information at the target timestamp. Second, Appearance Engram Activation maps the scaffold and reference frame into a shared latent space, where Latent Interaction Attention (LIA) enables structural tokens to activate relevant reference engrams throughout the denoising process of a one-step diffusion backbone. Furthermore, a skip connection with zero-initialized convolutions is incorporated to preserve structural information from the scaffold and suppress structure drift during generation. Following the standard one-step restoration paradigm, we adapt pretrained SD-Turbo~\cite{sauer2025adversarial} with LoRA~\cite{hu2022lora} and perform a single denoising pass. The restored scaffold latent is then decoded into the target frame.

\begin{figure*}[t]
\centering
\includegraphics[width=\textwidth]{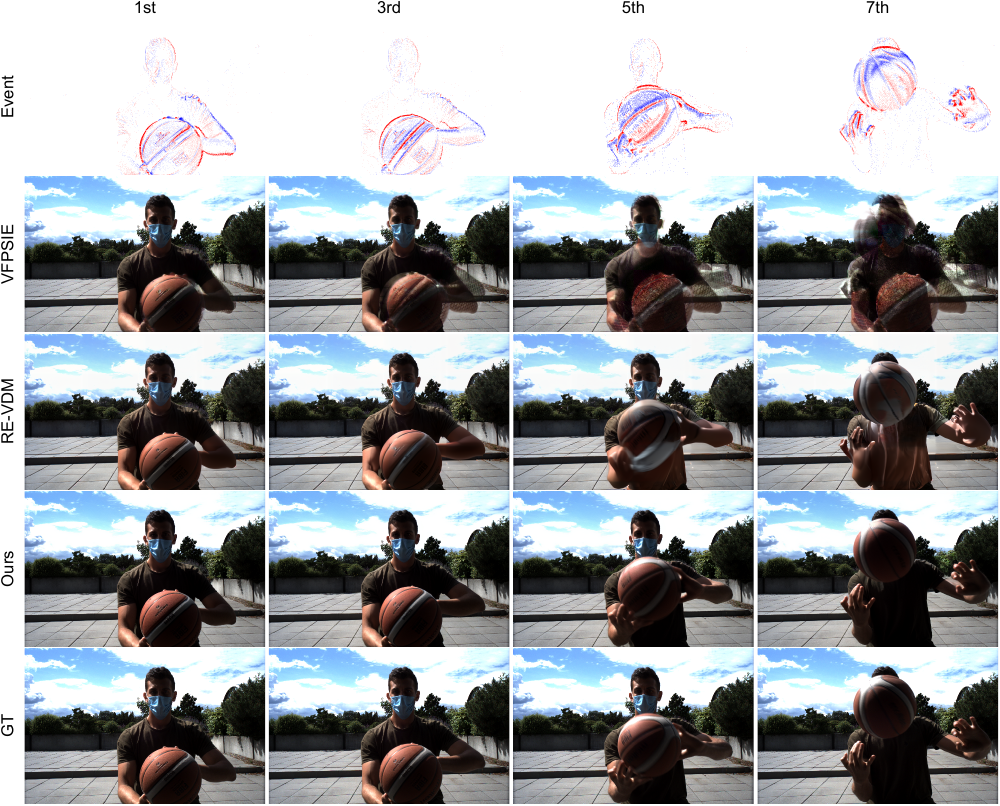}
\caption{Qualitative comparison on the BS-ERGB dataset. Columns show reconstructed target frames at the 1-, 3-, 5-, and 7-skip intervals. Rows show the event visualization, VFPSIE, RE-VDM, our method, and the ground truth.}
\label{fig:bsergb_qualitative}
\end{figure*}

\subsection{Structural Scaffold Estimation}
To capture the motion cues conveyed by asynchronous events, we convert the event stream between the reference and target timestamps into a spatiotemporal event voxel representation, as illustrated in Fig.~\ref{fig:framework}(b).

Event cameras asynchronously output events represented as tuples $e=(x,y,t,p)$, where $(x,y)$ denotes the spatial coordinates in the image plane, $t$ is the timestamp, and $p \in \{-1, +1\}$ indicates the polarity of the log-intensity change. Given a reference RGB frame $I_{t_0}$ and the subsequent event stream $E_{t_0 \rightarrow t_k}$, reference-based event-to-video reconstruction aims to recover the target RGB frame $I_{t_k}$.
Given $E_{t_0 \rightarrow t_k}=\{e_i\}_{i=1}^{N_e}$, we uniformly divide $[t_0,t_k]$ into $N$ temporal bins and map each event timestamp $t_i$ to a continuous temporal coordinate:

\begin{equation}
\tau_i=(N-1)\frac{t_i-t_0}{t_k-t_0}.
\end{equation}

To avoid assigning each event to a single bin, we distribute it across neighboring temporal bins through linear interpolation while retaining its spatial location $(x_i,y_i)$ and polarity $p_i$. Accordingly, the $b$-th slice of the event voxel is computed as

\begin{equation}
V_b(x,y)=\sum_{i\in S_{x,y}} p_i \max(0,1-|b-\tau_i|),
\end{equation}

where $b=0,\ldots,N-1$ denotes the temporal bin index, and $S_{x,y}$ is the set of indices of events occurring at pixel $(x,y)$. The linear interpolation kernel assigns each event to at most two adjacent temporal bins. The resulting event voxel $V$ has dimensions $N \times H \times W$.

After constructing the event voxel, we estimate a motion-structure scaffold at the target timestamp using a lightweight dual-encoder U-Net. The event voxel $V$ and reference frame $I_{t_0}$ are processed by separate encoders, and their features are concatenated at corresponding levels and decoded into $\tilde{I}_{t_k}^{s}$. The scaffold is not required to be photorealistic; its primary role is to expose motion boundaries and scene structures to the diffusion backbone.

\subsection{Appearance Engram Activation}
Given the motion-structure scaffold $\tilde{I}_{t_k}^{s}$ and reference frame $I_{t_0}$, a frozen VAE encoder maps them into latent representations $z_s$ and $z_r$, respectively:
\begin{equation} 
z_{s}=\mathcal{E}(\tilde{I}_{t_k}^{s}),
\qquad 
z_{r}=\mathcal{E}(I_{t_0}). 
\end{equation}

The tokens in $z_s$ encode target-time structural information, whereas those in $z_r$ serve as appearance engrams that retain rich visual information from the reference frame. To associate scaffold-derived structural tokens with relevant appearance engrams, we instantiate Latent Interaction Attention (LIA). Specifically, the scaffold and reference latents are stacked along a new source dimension to form
$z\in \mathbb{R}^{B\times M\times C\times H\times W}$,
where $B$ denotes the batch size, $M=2$ corresponds to the scaffold and reference latents, and $C$, $H$, and $W$ denote the channel and spatial dimensions.

\begin{table*}[t]
\centering
\footnotesize
\setlength{\tabcolsep}{2.0pt}
\renewcommand{\arraystretch}{1.12}

\caption{Quantitative comparison on three event-based video benchmarks, averaged over the indicated skip ranges. Bold denotes the best performance under the matched-input setting $I_0+\mathrm{Events}$; gray rows denote bidirectional interpolation methods.}
\label{tab:quantitative_comparison}

\begin{tabularx}{0.985\textwidth}{@{}l|c|c|YYY|YYY|YYY@{}}
\specialrule{1.0pt}{0pt}{0pt}

\multirow{2}{*}{Method} &
\multirow{2}{*}{Category} &
\multirow{2}{*}{Input} &
\multicolumn{3}{c|}{\rule{0pt}{2.6ex}BS-ERGB (1--7 skips)} &
\multicolumn{3}{c|}{\rule{0pt}{2.6ex}ERF-X170FPS (1--15 skips)} &
\multicolumn{3}{c}{\rule{0pt}{2.6ex}HQ-EVFI (1--31 skips)} \\

\cline{4-12}

& & &
\rule{0pt}{2.6ex}PSNR$\uparrow$ &
\rule{0pt}{2.6ex}SSIM$\uparrow$ &
\rule{0pt}{2.6ex}LPIPS$\downarrow$ &
\rule{0pt}{2.6ex}PSNR$\uparrow$ &
\rule{0pt}{2.6ex}SSIM$\uparrow$ &
\rule{0pt}{2.6ex}LPIPS$\downarrow$ &
\rule{0pt}{2.6ex}PSNR$\uparrow$ &
\rule{0pt}{2.6ex}SSIM$\uparrow$ &
\rule{0pt}{2.6ex}LPIPS$\downarrow$ \\

\hline

DMVFN
& \multirow{2}{*}{\makecell[c]{Frame\\Prediction}}
& \multirow{2}{*}{RGB frames}
& 19.58 & 0.65 & 0.27
& 19.10 & 0.66 & 0.32
& 19.22 & 0.71 & 0.22 \\

MGU
& &
& 18.97 & 0.59 & 0.40
& 17.22 & 0.65 & 0.55
& 17.61 & 0.70 & 0.29 \\

\hline

CISTA-LSTC
& \multirow{2}{*}{\makecell[c]{Event-only\\Reconstruction}}
& \multirow{2}{*}{Events}
& 8.00 & 0.36 & 0.70
& 10.29 & 0.47 & 0.64
& 10.38 & 0.52 & 0.53 \\

HyperE2VID
& &
& 10.09 & 0.34 & 0.61
& 10.24 & 0.40 & 0.78
& 11.85 & 0.37 & 0.63 \\

\hline

VFPSIE
& \multirow{3}{*}{\makecell[c]{Reference-based\\Reconstruction}}
& \multirow{3}{*}{$I_0 + \mathrm{Events}$}
& 22.00 & 0.69 & 0.14
& 20.67 & 0.63 & 0.24
& 21.77 & 0.75 & 0.17 \\

RE-VDM
& &
& 23.62 & 0.73 & 0.13
& 21.25 & 0.70 & 0.21
& 22.80 & 0.76 & 0.15 \\

\textbf{Ours}
& &
& \textbf{25.08} & \textbf{0.76} & \textbf{0.08}
& \textbf{24.54} & \textbf{0.75} & \textbf{0.14}
& \textbf{24.90} & \textbf{0.80} & \textbf{0.07} \\

\hline

\rowcolor{gray!12}
CBMNet
&
&
& 25.47 & 0.77 & 0.11
& 26.85 & 0.81 & 0.09
& 27.61 & 0.83 & 0.09 \\

\rowcolor{gray!12}
TimeLens-XL
& \multirow{-2}{*}{\makecell[c]{\textbf{Bidirectional}\\\textbf{Interpolation}}}
& \multirow{-2}{*}{$I_0 + I_T + \mathrm{Events}$}
& 26.28 & 0.78 & 0.07
& 27.23 & 0.82 & 0.06
& 28.08 & 0.88 & 0.03 \\

\specialrule{1.0pt}{0pt}{0pt}

\end{tabularx}
\end{table*}

Within each transformer block, LIA rearranges the source and spatial dimensions into a joint token sequence, allowing structural tokens and appearance engrams to interact through the self-attention layers. The resulting tokens are then rearranged back to the original latent layout and passed to subsequent layers:
\begin{equation}
\begin{alignedat}{2}
&\bar{z}
&&=
\operatorname{Rearrange}
\left(
z,\; B \times (MHW) \times C
\right)
\\
&\hat{z}
&&=
\operatorname{Self\text{-}Attention}(\bar{z})
\\
&z'
&&=
\operatorname{Rearrange}
\left(
\hat{z},\; B \times M \times C \times H \times W
\right)
\end{alignedat}
\end{equation}

Through joint latent interaction, LIA enables scaffold tokens to activate relevant reference engrams. Because LIA reuses pretrained self-attention layers, it adds no extra parameters and supports parameter-efficient adaptation with LoRA. During denoising, the scaffold-derived structural tokens serve as target-time queries, while the reference branch provides an appearance engram bank; only the updated scaffold latent is decoded. This token-space association does not require fixed pixel-wise correspondence, allowing relevant reference information to remain accessible under large motion and longer temporal intervals, while the diffusion prior complements uncertain or newly visible regions.

To apply the fixed-resolution diffusion backbone to high-resolution inputs, we partition the scaffold and reference latent token maps, $z_s$ and $z_r$, into overlapping tiles of size $S \times S$ with stride $r$. Tile-wise LIA and one-step denoising then produce the restored scaffold tiles $\tilde{z}_j$, which are merged through normalized weighted averaging:
\begin{equation}
\tilde{z} =
\frac{\sum_j P_j(W \odot \tilde{z}_j)}
{\sum_j P_j(W)},
\end{equation}
where $P_j(\cdot)$ places the $j$-th tile on the full latent canvas, $W$ is a cosine weighting window that smooths overlapping boundaries, and $\odot$ denotes element-wise multiplication. The merged latent $\tilde{z}$ is decoded into the high-resolution target frame.

\section{Experiments}

\begin{figure}[t]
\centering
\includegraphics[width=\columnwidth]{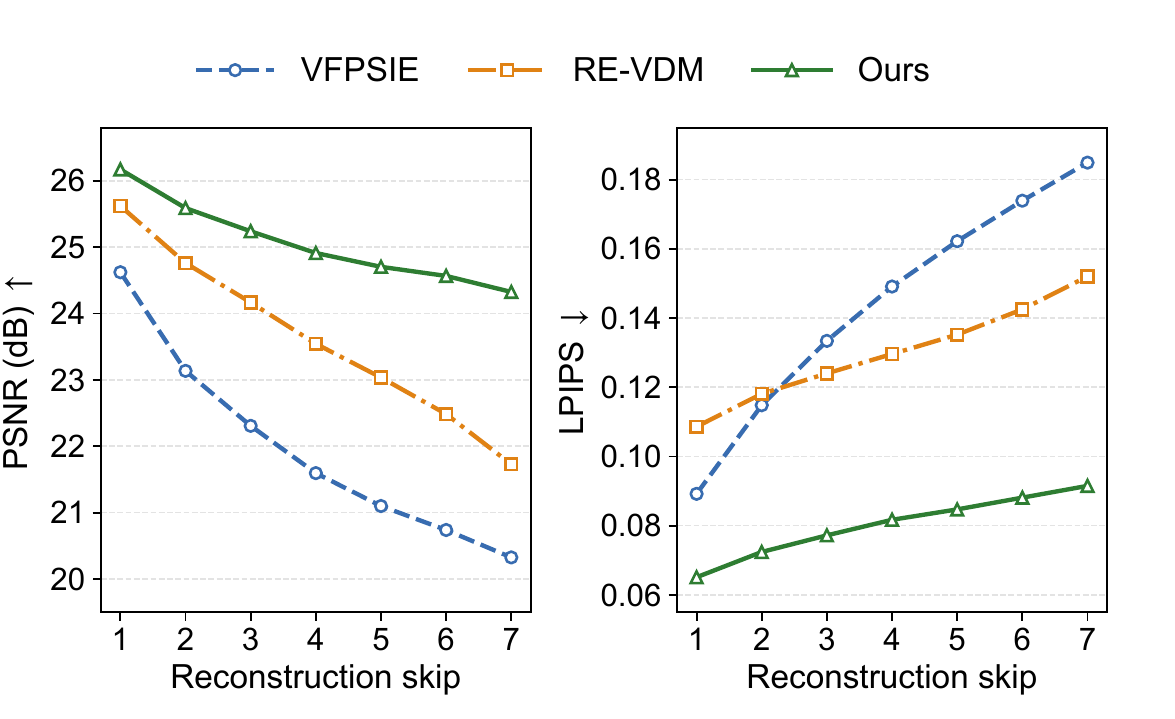}
\caption{Performance across different reconstruction skips on the BS-ERGB dataset. All methods use the same input setting with a single reference frame and subsequent events.}
\label{fig:skip_curve}
\end{figure}

\begin{figure*}[t]
\centering
\includegraphics[width=\textwidth]{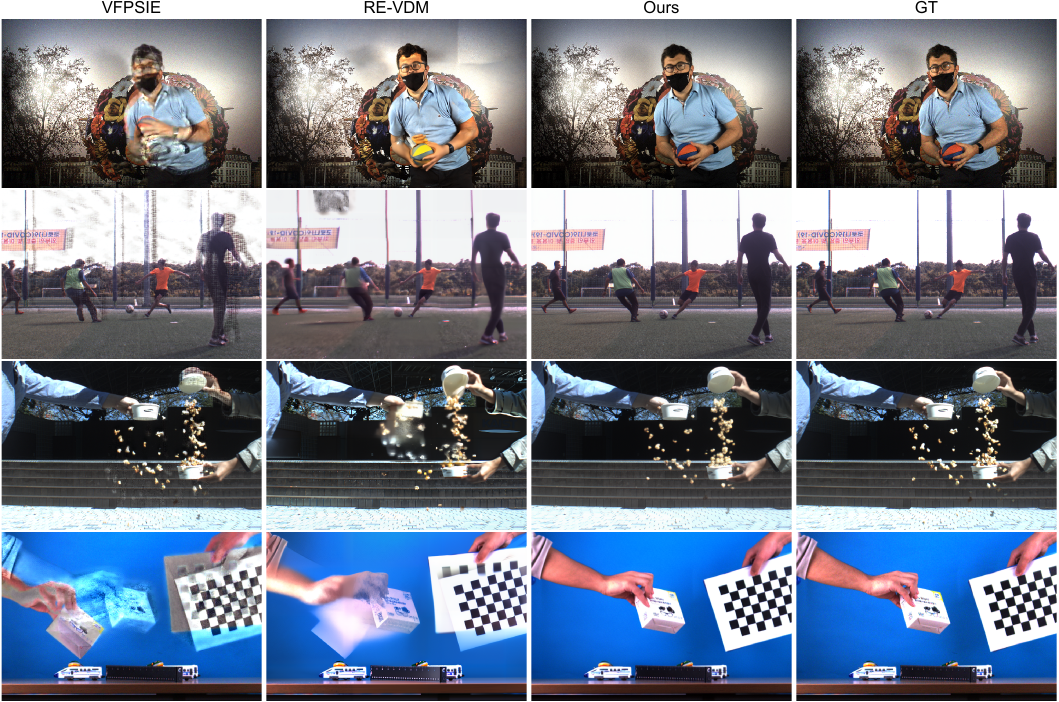}
\caption{Qualitative comparison on BS-ERGB, ERF-X170FPS, and HQ-EVFI at their respective maximum skip settings. Columns show the reconstruction results of VFPSIE, RE-VDM, our method, and the ground truth.}
\label{fig:cross_dataset_qualitative}
\end{figure*}

\subsection{Experimental Setup}
\subsubsection{Dataset Preparation}

We evaluate our method on three real-world synchronized RGB-event datasets: BS-ERGB~\cite{Tulyakov_2022_CVPR}, ERF-X170FPS~\cite{Kim_2023_CVPR}, and HQ-EVFI~\cite{ma2025timelensxl}. For each sequence, we select an RGB frame at timestamp $t_0$ as the reference and a subsequent timestamp $t_k$ as the reconstruction target. The reference frame $I_{t_0}$ and events over $[t_0,t_k]$ form the model input, while the RGB frame $I_{t_k}$ serves as ground truth. Varying $t_k$ yields reconstruction samples with different reference-target intervals.
We train a separate model for each dataset using its official training split and evaluate it on the corresponding test split. All results are reported at the native image resolution.

\subsubsection{Implementation Details}

The structural scaffold estimator is implemented as a lightweight dual-encoder U-Net with four scales and 32 base channels. Each event stream is represented as a voxel grid with $N=20$ temporal bins, and the network produces a three-channel scaffold at the target timestamp. To emphasize motion-active regions during training, we employ an event-weighted Charbonnier loss:
\begin{equation}
\mathcal{L}_s=
\frac{1}{|\Omega|}
\left\lVert
(\lambda_b+\lambda_e M_e)
\odot
\rho(\tilde{I}_{t_k}^{s}-I_{t_k}^{gt})
\right\rVert_1,
\end{equation}
where $\tilde{I}_{t_k}^{s}$ is the estimated scaffold, $\rho(\cdot)$ denotes the element-wise Charbonnier penalty, $M_e$ is the normalized event-activity map, and $\Omega$ is the image domain. We set $\lambda_b=0.2$ and $\lambda_e=1.8$ to assign larger weights to motion-active regions. The estimator is trained for 50 epochs on a single NVIDIA A800 GPU using AdamW with a learning rate of $1\times10^{-4}$.

For the diffusion stage, we fine-tune SD-Turbo~\cite{sauer2025adversarial} under its standard one-step restoration setting. The estimated scaffold and reference RGB frame jointly condition the reconstruction, while the target RGB frame provides supervision. We optimize the model using a pixel-wise reconstruction loss and an LPIPS perceptual loss~\cite{Zhang_2018_CVPR}:

\begin{equation}
\mathcal{L}_d=\mathcal{L}_2+\lambda\mathcal{L}_{\mathrm{LPIPS}},
\end{equation}

where $\mathcal{L}_2$ and $\mathcal{L}_{\mathrm{LPIPS}}$ denote the pixel-wise and perceptual losses, respectively, and $\lambda=1$. The model is trained on a single NVIDIA A800 GPU using Adam with a learning rate of $2\times10^{-5}$ and a batch size of 4. We apply LoRA to the diffusion U-Net and VAE decoder with ranks of 16 and 4, respectively. For high-resolution inference, we partition the scaffold and reference latent token maps into $64\times64$ tiles ($S=64$) with a stride of $r=48$. This tile-wise procedure is used only during inference.

\subsubsection{Evaluation Strategy}
To comprehensively evaluate reference-based event-to-video reconstruction, we compare Engram-E2VID with representative methods under different input settings, as summarized in Table~\ref{tab:quantitative_comparison}. The compared methods are grouped into four categories: (1) RGB-only frame prediction methods, which predict target frames from multiple preceding RGB frames without using events, i.e., DMVFN~\cite{10204820} and MGU~\cite{zhong2024motion}; (2) event-only reconstruction methods, which recover intensity frames directly from event streams without a reference image, i.e., CISTA-LSTC~\cite{10130595} and HyperE2VID~\cite{10462903}; (3) reference-based reconstruction methods, which use a reference RGB frame and subsequent events to reconstruct target frames, i.e., VFPSIE~\cite{zhu2024video} and RE-VDM~\cite{Chen_2025_CVPR}; and (4) bidirectional event-based interpolation methods, which use two endpoint RGB frames together with the intervening event stream, i.e., CBMNet~\cite{Kim_2023_CVPR} and TimeLens-XL~\cite{ma2025timelensxl}. For RE-VDM, which supports both interpolation and generation, we adopt its official single-reference generation mode using the initial frame and subsequent events. The third category provides the matched-input comparison for Engram-E2VID. In contrast, the fourth category has access to an additional future endpoint frame and therefore enjoys an information advantage; we report these results separately as privileged-input references rather than direct comparisons.

For the reconstructed frames, we report PSNR, SSIM~\cite{1284395}, and LPIPS as evaluation metrics. To evaluate reconstruction across different temporal intervals, we adopt a dataset-specific maximum skip $K$, taking into account the frame rate and temporal characteristics of each dataset. For each reference frame, target frames are reconstructed at all skip values from $1$ to $K$. We first compute the performance at each skip and then average the skip-wise results to obtain the overall score, ensuring equal contribution from different temporal intervals. Specifically, we set $K=7$ for BS-ERGB, $K=15$ for ERF-X170FPS, and $K=31$ for HQ-EVFI. Larger skip values correspond to longer temporal gaps between the reference and target frames, generally involving greater motion displacement and increased reconstruction difficulty.

\subsection{Evaluation on Reference-Based Reconstruction}

The quantitative results are reported in Table~\ref{tab:quantitative_comparison}. Engram-E2VID substantially outperforms RGB-only prediction and event-only reconstruction methods, demonstrating the benefit of jointly leveraging appearance information from the reference frame and fine-grained temporal cues from events. Among reference-based reconstruction methods with the same input configuration, Engram-E2VID consistently achieves the best performance across all three benchmarks. Compared with the strongest same-input baseline, RE-VDM, our method improves PSNR by 1.46, 3.29, and 2.10 dB on BS-ERGB, ERF-X170FPS, and HQ-EVFI, respectively. Although bidirectional interpolation methods benefit from an additional endpoint frame and generally achieve higher PSNR and SSIM, Engram-E2VID remains competitive in perceptual quality. In particular, it achieves lower LPIPS than CBMNet on BS-ERGB (0.08 vs. 0.11) and HQ-EVFI (0.07 vs. 0.09), highlighting the effectiveness of the proposed reconstruction framework.

Figure~\ref{fig:bsergb_qualitative} compares reconstruction results on BS-ERGB under the 1-, 3-, 5-, and 7-skip settings. As the temporal interval increases, conventional reference-based methods such as VFPSIE exhibit increasingly noticeable blur, ghosting, and structural artifacts. RE-VDM partially alleviates these degradations through generative priors, but may still produce color inconsistencies, missing details, or visually implausible content at larger temporal intervals. In contrast, Engram-E2VID better preserves object boundaries, appearance details, and overall scene structure across different skip settings. \mbox{Figure~\ref{fig:cross_dataset_qualitative}} further compares representative scenes across all three datasets at their respective maximum evaluated skips, spanning diverse object scales and viewing distances. Engram-E2VID consistently retains sharper structures and more faithful appearance.

To further assess robustness to increasing reference-target intervals, we analyze the variation of PSNR and LPIPS with reconstruction skip on BS-ERGB, as illustrated in Fig.~\ref{fig:skip_curve}. All methods in this analysis use the same input setting of a single reference frame and subsequent events. As the reconstruction skip increases from 1 to 7, Engram-E2VID consistently outperforms VFPSIE and RE-VDM while exhibiting slower performance degradation. Specifically, its PSNR decreases by only 1.85 dB, compared with 4.30 dB for VFPSIE and 3.89 dB for RE-VDM. Its LPIPS increases by only 0.026, compared with 0.096 for VFPSIE and 0.043 for RE-VDM. This substantially slower degradation provides direct evidence of the long-interval reconstruction advantage of Engram-E2VID. As the reference-target interval increases, token-space association enables target-time structural tokens to access relevant appearance engrams despite growing spatial displacement, without requiring direct pixel-wise correspondence. The diffusion prior further complements uncertain or newly revealed content, allowing Engram-E2VID to preserve reconstruction fidelity over longer intervals.

\begin{figure}[t]
\centering
\includegraphics[width=\columnwidth]{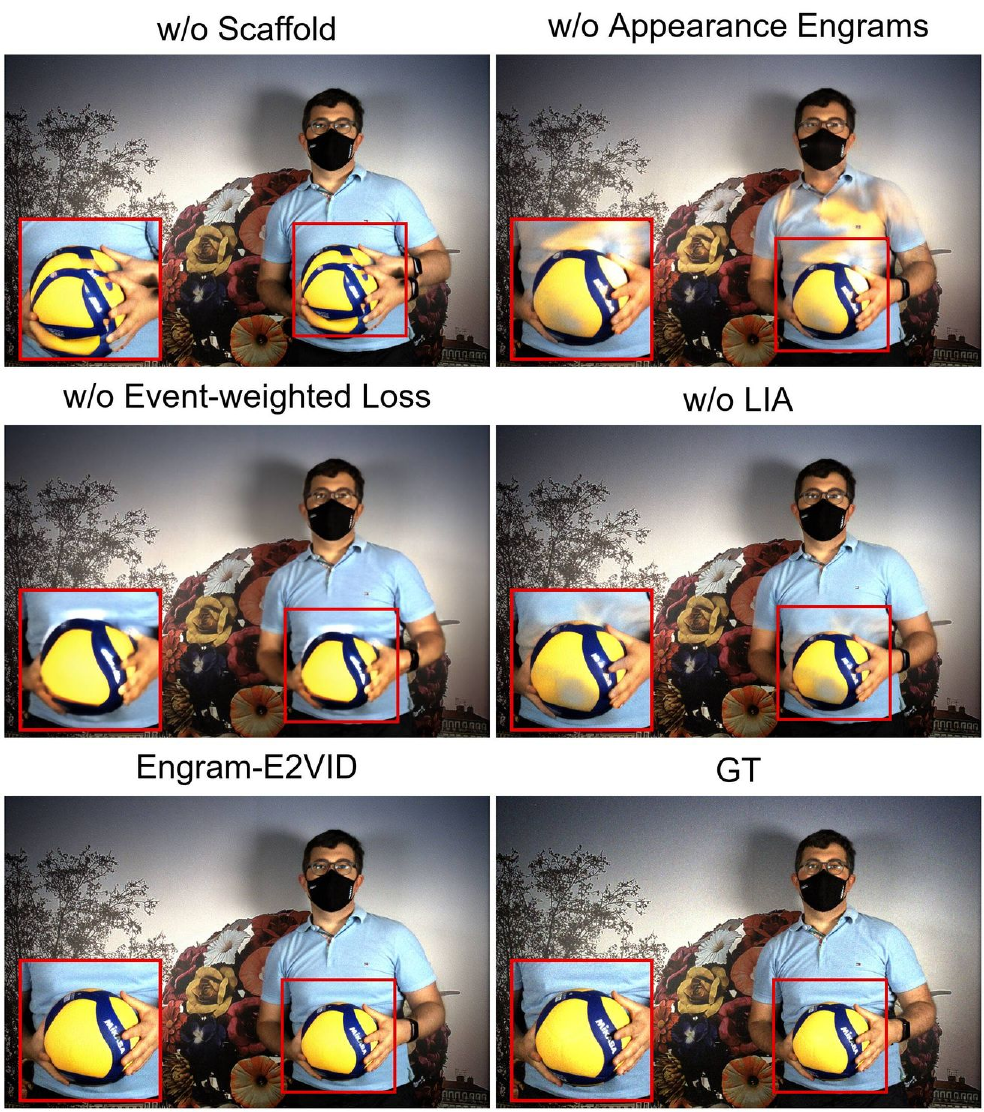}
\caption{Qualitative ablation study on BS-ERGB at the 7-skip setting. The first two rows show the ablated variants, while the last row shows the full model and ground truth.}
\label{fig:ablation}
\end{figure}

\subsection{Ablation Studies}

We conduct ablation studies to investigate the effects of key components in Engram-E2VID, with the quantitative results reported in Table~\ref{tab:ablation}. In the w/o Scaffold variant, we remove the scaffold estimator and collapse the event voxel into a normalized three-channel event image as the structural input to the diffusion backbone. This replacement causes the largest performance degradation, reducing PSNR from 25.08 dB to 21.86 dB. The substantial drop validates the need to transform event observations into explicit target-time structural guidance before diffusion restoration. Using a standard Charbonnier loss instead of the proposed event-weighted variant further decreases PSNR by 0.46 dB and degrades both SSIM and LPIPS, showing that emphasizing motion-active regions provides more effective supervision for scaffold estimation.

We further ablate appearance engrams and LIA. For the w/o Appearance Engrams variant, we replace the reference-frame input to the Appearance Engram Activation stage with the estimated scaffold, so that both branches receive the scaffold. This decreases PSNR by 2.34 dB and increases LPIPS from 0.0856 to 0.1842, confirming the importance of appearance engrams. In the w/o LIA variant, the scaffold and engram tokens are processed with separate self-attention, preventing token-level information exchange between them. This reduces PSNR by 0.67 dB and degrades all three metrics. As shown in Fig.~\ref{fig:ablation}, the ablated variants exhibit visible degradation in motion-active regions, including blurred boundaries and loss of fine details around the moving ball and hands, while the full model better preserves local structure and appearance.

\begin{table}[t]
\centering
\small
\setlength{\tabcolsep}{7pt}
\renewcommand{\arraystretch}{1.15}
\caption{Ablation study of key components in Engram-E2VID on BS-ERGB, averaged over reconstruction skips from 1 to 7.}
\label{tab:ablation}

\begin{tabular}{lccc}
\toprule
\textbf{Variant}
& \textbf{PSNR}$\uparrow$
& \textbf{SSIM}$\uparrow$
& \textbf{LPIPS}$\downarrow$ \\
\midrule

w/o Scaffold
& 21.86 & 0.6521 & 0.2176 \\

w/o Appearance Engrams
& 22.74 & 0.6817 & 0.1842 \\

w/o Event-weighted Loss
& 24.62 & 0.7418 & 0.1035 \\

w/o LIA
& 24.41 & 0.7324 & 0.1124 \\

\midrule

\textbf{Engram-E2VID}
& \textbf{25.08}
& \textbf{0.7622}
& \textbf{0.0856} \\

\bottomrule
\end{tabular}
\end{table}

\section{Conclusion}
We present Engram-E2VID, a reference-based event-to-video reconstruction framework that transforms events into a target-time structural scaffold and organizes reference appearance as token-space engrams. Within a one-step diffusion backbone, structural tokens activate relevant appearance engrams, enabling faithful reconstruction under complex motion and increasing temporal intervals. Experiments across three benchmarks demonstrate superior reconstruction quality and slower degradation as the temporal interval increases. Qualitative results further indicate that independently reconstructed target frames remain temporally coherent despite the absence of explicit temporal consistency regularization.

{
    \small
    \printreferences
}

\end{document}